\documentclass[a4paper,final,12pt,onecolumn]{IEEEtran}
\usepackage{setspace}
   \usepackage[dvips]{graphicx}
   \graphicspath{{../eps/}}
   \DeclareGraphicsExtensions{.pdf,.jpeg,.png,.eps}
\usepackage[cmex10]{amsmath}
\usepackage[tight,footnotesize]{subfigure}

\begin{document}
%
% paper title
% can use linebreaks \\ within to get better formatting as desired
\title{Memristor Crossbar-based Hardware Implementation of IDS Method}
%
%
% author names and IEEE memberships
% note positions of commas and nonbreaking spaces ( ~ ) LaTeX will not break
% a structure at a ~ so this keeps an author's name from being broken across
% two lines.
% use \thanks{} to gain access to the first footnote area
% a separate \thanks must be used for each paragraph as LaTeX2e's \thanks
% was not built to handle multiple paragraphs
%

\author{Farnood~Merrikh-Bayat,
        Saeed Bagheri-Shouraki, and Ali Rohani}% <-this % stops a space
\maketitle

\begin{abstract}
Ink Drop Spread (IDS) is the engine of Active Learning Method (ALM),
which is the methodology of soft computing. IDS, as a pattern-based
processing unit, extracts useful information from a system subjected
to modeling. In spite of its excellent potential in solving problems
such as classification and modeling compared to other soft computing
tools, finding its simple and fast hardware implementation is still
a challenge. This paper describes a new hardware implementation of
IDS method based on the memristor crossbar structure. In addition of
simplicity, being completely real-time, having low latency and the
ability to continue working after the occurrence of power breakdown
are some of the advantages of our proposed circuit.

\end{abstract}
%
%\keywords{show-through reduction, non-negative matrix factorization,
%blurring effect and nonlinear modeling}

\section{Introduction}
\label{intro}

Humans possess an excellent ability to process complicated
information with ease while their processing by using classical
mathematical approaches is very difficult if it is not impossible.
In other words, using very exact mathematical techniques in the
calculation of inferencing procedures seems to be in contrast with
human natural abilities. Where the most prominent specifications of
the mind are extracting and using pure qualitative knowledge,
approaching the real brain abilities requires a real soft computing
method executable on a non-exact calculating instrument with
uncertainty.

By considering the way which human being looks at any phenomenon,
acquires data from it and extracts knowledge about the observed
events, and the decision he makes to behave properly, Shouraki {\it
et al.} in \cite{shouraki1,shouraki2} proposed a fuzzy modeling
technique called ALM.

Processing nature of ALM is similar to that of neural network while
its resemblance to fuzzy logic is due to the fact that both are
modeled based on information-handling processes of the human brain.
ALM is mostly characterized by its intuitive pattern-based
processing which is based on the hypothesis that humans interpret
information in the form of pattern-like images rather than numerical
or logical forms.

ALM inference system consists of several pattern processing engines
called IDS which their rule is to extract useful information from
the system subjected to modeling. Murakami et al. in \cite{murakami}
have shown that IDS method has excellent potential in solving
problems such as classification and modeling compared to other soft
computing tools. In addition, they have reported that IDS method
possesses excellent real-time capabilities \cite{murakami2}.
However, their proposed digital system as a hardware implementation
of IDS method \cite{murakami2} or the one suggested by Tarkhan {\it
et al.} \cite{tarkhan} is so much complicated than the one predicted
and expected by Shouraki \cite{shouraki1}. In fact, one of the major
concerns during the proposition of ALM was the simplicity of its
hardware implementation. Lack of such a hardware is the main reason
which causes ALM not to become popular like other modeling
approaches. As a result, this paper describes memristor
crossbar-based hardware implementation of basic IDS unit with the
ability of extracting narrow path and spread. In addition of
simplicity, being completely real-time, having low latency and the
ability to continue working after the occurrence of power breakdown
are some of the advantages of our proposed circuit.

The paper is organized as follows. IDS method is briefly described
in Section \ref{IDS}. Section \ref{memristor} is devoted to the
explanation of the newly found fourth circuit element called
memristor. Crossbar-based hardware implementation of IDS method is
presented in Section \ref{main}. Eventually, a few experimental
results are presented in Section \ref{simulation}, before
conclusions in Section \ref{conclusion}.

\section{Brief description of IDS method}
\label{IDS}

IDS method works by breaking down a complex system into simpler
parts similar to the manner in which humans behave confronting
complicated subjects. For multi-input single-output (MISO) system,
this can be done simply by splitting the MISO system, $y=f(x_1, x_2,
\ldots, x_N)$, into multiple single-input single-output (SISO) ones.
From available input-output training data, each SISO constructs a
pattern which will then be used in modeling process in IDS method.

\begin{figure}[!t]
\centering %\subfigure[Input recto image with show-through.]
{
% label for first subfigure
\fbox{\includegraphics[width=3.3in,height=1.9in]{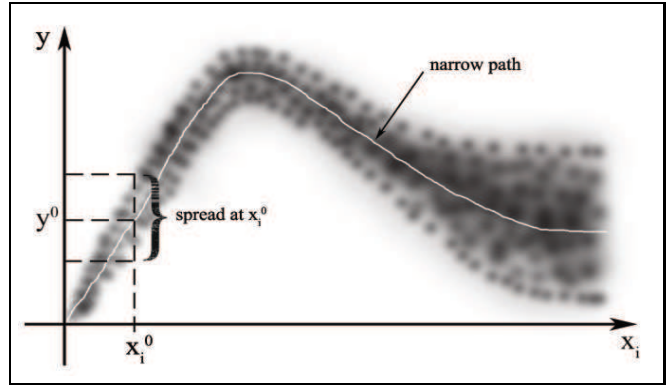}}}
\caption{Typical plane shows data spreading and extraction of narrow
path and spread.}
\label{fig9} %% label for entire figure
\end{figure}

For any MISO system, IDS method considers N 2-dimensional discrete
planes which N is the number of inputs in this MISO system and in
the {\it i}th plane, vertical and horizontal axes are y and $x_i$
respectively. Then, each of the available input-output training data
is spread on all of these planes such as the one shown in Fig.
\ref{fig9} which shows one of these planes after data spreading.
This process is called ``data spread'' or ``distilling ink drop''
since it seems that at the coordinate specified by each of the
input-output data, one ink drop has been distilled on each of these
planes. As individual data spreads overlap each other, the
overlapping regions become increasingly darker finally resulting in
a pattern on each of the planes. By applying IDS method to
constructed pattern image of each plane, two different types of
information would be extracted. One is the narrow path and the other
is the deviation of the spread data points around each narrow path
(see Fig. \ref{fig9}). Narrow path in the {\it i}th plane shows the
overall relationship between the output of the system, y, and the
{\it i}th input $x_i$ while the spread of the data points around
this path shows the importance degree of $x_i$ in system behavior.
For example, if the spread of data points around narrow path be
sufficiently small in the $x_1-y$ plane compared to other planes, it
means that the importance degree of the input $x_1$ is higher than
any other inputs. In other word, in this case MISO system can be
approximately simplified to the SISO system described by the narrow
path in the $x_1$-y plane.

At the arrival of any new input vector $\mathbf{x}^t=[x_1^t,
x_2^t,\ldots , x_N^t]$, for determining the corresponding output by
inference, values of narrow paths and spreads should be calculated
at this point from the pattern images of the planes. These values
are then transferred from IDS units to the upper layer for being
used for the inferential process in the ALM. ALM uses these
information and approximates output value. Here, we do not concern
with how ALM performs inference, instead, we will only concentrate
on hardware implementation of data spreading and extraction of
narrow path and spread values.

Different schemes can be defined for extracting narrow path and
spread from the pattern image in each of the planes. Hereafter, in
order to have simple hardware implementation and not involving in
exact mathematical techniques, narrow path and spread are computed
as described below. Assume that a new input data
$\mathbf{x}^0=[x_1^0, x_2^0,\ldots , x_N^0]$ is observed and we want
to calculate values of narrow paths and spreads from the pattern
images of planes at this input point. Without loss of generality,
consider the pattern image after data spreading in $x_1-y$ plane
(such as the one shown in Fig. \ref{fig9}). The number of cells in
the column of this plane correspond to $x_1^0$ which have ink more
than the predetermined value can be considered as a spread value at
this point. Moreover, $y^0$ will be the value of narrow path at this
point provided that the sum of the ink of the cells above $y^0$ be
equal to the sum of the ink of the cells below it. This procedure
can be repeated for each of the planes for extracting values of
narrow paths and spreads at input point $\mathbf{x}^0$.

Using these definitions for narrow path and spread, in Section
\ref{main} we present a new memristor crossbar-based structure as a
hardware implementation of IDS method.

\section{Memristor}
\label{memristor}

 Publication of a paper \cite{williams} in Nature
by Hewlett Packard labs in May 1, 2008, which announced the first
experimental realization of the memristor whose existence was
predicted in 1971 by Leon Chua \cite{Chua} has caused an
extraordinary increased interest in this passive circuit element. It
has become clear that memristor can have many potential applications
such as non-volatile memory construction \cite{waser}, creation of
analog neural network and emulation of human learning
\cite{pershin}, building programmable analog circuits \cite{Shin}
and in signal processing field \cite{Mouttet1, Mouttet2, Mouttet3}.

Memristor ($M$), different from other electrical parameters namely
resistance, capacitance and inductance, denotes the relationship
between magnetic flux($\varphi$) and electric charge ($q$) as
\cite{Chua}:
\begin{equation}
d\varphi=Mdq.
\end{equation}

By rewriting this equation, memristance can be expressed as:

\begin{equation}
M(q)=\frac{d\varphi/dt}{dq/dt}=\frac{V(t)}{I(t)},
\end{equation}
which shows that the unit of memristance is ohm. A memristor can be
thought of as a resistive device that its resistance varies in
dependence of its current or magnetic flux.

\begin{figure}[!t]
\centering %\subfigure[Input recto image with show-through.]
{
% label for first subfigure
\fbox{\includegraphics[width=3in,height=1.3in]{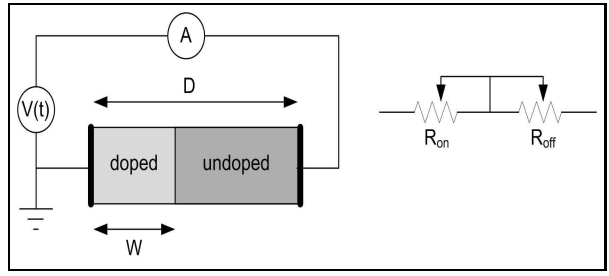}}}
\caption{Structure of the memristor reported by HP researchers and
its equivalent circuit model.}
\label{fig8} %% label for entire figure
\end{figure}

Memristor is an electrically switchable semiconductor thin film
sandwiched between two metal contact with a total length of $D$ and
consists of doped and un-doped regions which its physical structure
with its equivalent circuit model is shown in Fig.~\ref{fig8}
\cite{williams}. The internal state variable $w$ determines the
length of doped region with low resistance against un-doped region
with high resistivity. This internal state variable and consequently
the total resistivity of the device can be changed by applying
external voltage bias $v(t)$ \cite{williams2}. If the doped region
extends to the full length $D$, the total resistance of the device
will be at its lowest level denoted as $R_{on}$ and If the un-doped
region extends to the full length $D$, the total resistance of the
device will be at its highest level namely $R_{off}$. Thus, the
mathematical model for the total resistance of the memristor can be
written as \cite{williams}:
\begin{eqnarray}\label{eq15}
% \nonumber to remove numbering (before each equation)
  M(w)&=&R_{on}\frac{w}{D}+R_{off}\left(1-\frac{w}{D}\right),\nonumber \\
  w(t)&=&w_0+\frac{\mu_vR_{on}}{D}q(t),
\end{eqnarray}
%\begin{equation}\label{eq15}
%\begin{array}
%M(w)=R_{on}\frac{w}{D}+R_{off}\left(1-\frac{w}{D}\right)\nonumber\\
%w(t)=w_0+\frac{\mu_vR_{on}}{D}q(t),
%\end{array}
%\end{equation}
where $w_0$ is the initial state for state variable $w$, $\mu_v$ is
the average ion mobility and $q(t)$ is the amount of electric charge
(integral of current) that has passed through the device. Above
equations show that passing current from memristor in one direction
will increase the resistance while changing the direction of the
applied current will decrease its memristance. On the other hand, it
is obvious that in this element, passing current in one direction
for longer period of time (which means $q(t)$ has higher absolute
value) will change the resistance of the memristor more.

As a result, memristor is nothing else than the analog variable
resistor which its resistance can be adjusted by changing the
direction and duration of the applied voltage. Therefore, memristor
can be used as a storage device in which analog values can be stored
as an impedance instead of voltage.

\section{Memristor crossbar-based hardware implementation of IDS method}
\label{main}

\subsection{Using memristor crossbar for data spreading}
\label{main-1}
\begin{figure}[!t]
\centering %\subfigure[Input recto image with show-through.]
{
% label for first subfigure
\fbox{\includegraphics[width=3.3in,height=2.8in]{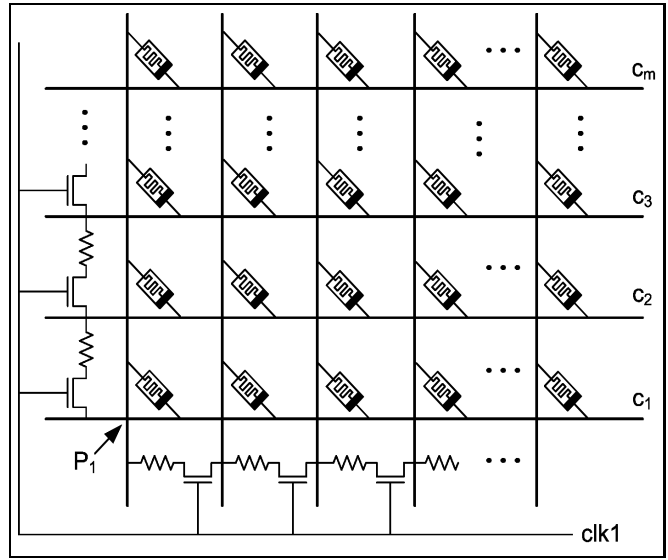}}}
\caption{A memristor crossbar designed for data spreading.}
\label{fig1} %% label for entire figure
\end{figure}

 Figure \ref{fig1} shows a memristor crossbar circuit
that we have proposed as a hardware implementation of data diffusing
process in IDS method. This memristor crossbar which is considered
as one of the N discrete 2-dimensional planes for data spreading
works as follows. This crossbar consists of two sets of horizontal
and vertical wires which each wire in one of these sets is
perpendicular to all of wires in the other set. The region where a
wire in one set crosses over a wire in the other set is called a
junction (or crosspoint). In this particular crossbar, memristor is
formed in each junction between two passing wires. In this case,
memristors in the crosspoints can be configured (their resistances
can be changed) by applying suitable voltages to these two
perpendicular wires. For example, memristance of the memristor
located at the crossing point of the first horizontal and the first
vertical wires of the crossbar correspond to coordinate (1, 1)
(denoted by $P_1$ in Fig. \ref{fig1}) can be increased by applying a
positive voltage to the first column of the memristor crossbar while
grounding the first horizontal wire. Dropping positive voltage
across the memristor will cause the current to flow through it which
will increase the memristance of this passive element. On the other
hand, reversing the polarity of the applied voltage will decrease
the memristance of the memristor. Moreover, as \eqref{eq15}
indicates, application of higher voltages for longer period of time
will increase the memristance of the memristor more. This means that
data can be stored in crosspoints through the memristances of the
memristors.

For utilizing this memristor crossbar structure for data spreading,
each wire should be connected to its predecessor and successor wires
by a simple resistor which can be done in Fig. \ref{fig1} by setting
{\it clk1} input pin into logic 1. As it will become clear soon,
resistance value of these resistors is inversely related to the
effecting radius of ink drops. Higher resistances will yield to the
lower effecting radius. Note that rows of the memristor crossbar
shown in Fig. \ref{fig1} are connected to the inputs of the circuit
shown in Fig. \ref{fig3}.

\begin{figure}[!t]
\centering %\subfigure[Input recto image with show-through.]
{
% label for first subfigure
\fbox{\includegraphics[width=2.5in,height=1.2in]{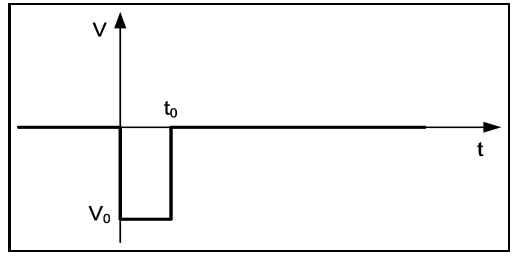}}}
\caption{A finite duration negative pulse used for ink dropping.}
\label{fig2} %% label for entire figure
\end{figure}

Suppose that initially all of the memristors in the crossbar have
their highest possible memristance value, {\it i.e.} $R_{off}$.
Dropping ink at any arbitrary point ($k_0$, $l_0$) can be
accomplished by grounding the horizontal wire corresponding to $l_0$
and applying a finite duration negative pulse (such as the typical
one shown in Fig. \ref{fig2}) to the vertical wire related to index
$k_0$. In this case, the current passing through the memristor at
point ($k_0$, $l_0$) will be maximum and hence, its resistance will
be decreased more than any other memristors in the crossbar. Since
all of rows or columns of the crossbar are connected to each others
with resistors, by applying this negative pulse, current will pass
through all of the memristors in the crossbar. However, by moving
away from coordinate ($k_0$, $l_0$) in each direction, the amount of
current flowing through the memristors will be reduced (because of
the resistors connected between columns or rows of the crossbar) and
therefore their memristances will be decreased less than those which
are nearer to the coordinate ($k_0$, $l_0$). Note that in order to
prevent unwanted feedback paths within the crossbar, a rectification
layer may be provided or p-type and n-type doping may be performed
creating diode junctions at each crosspoint. Next ink will be
dropped in the same manner. Distilling of each ink drop will
decrease the memristance of the memristors more (but with different
amounts depending to their distances from the dropping point) and by
this means, newly dropped ink will be added to the previously stored
ink pattern.

Here, it is worth to mention that since dropping each ink drop
alters memristance values of all of the memristors in the crossbar,
 changes caused in memristances by distilling two ink drops at the same
 coordinate ($k_0$, $l_0$) but in different times
will not be the same. This is because of the fact that passing
current from memristors in these two cases are not the same. For
relaxing this drawback, it is necessary that $R_{off}$ be much
higher than the maximum amount of changes that can be made in the
memristances of the memristors during ink dropping procedure. One of
the particular advantages of this structure is that stored pattern
image (as the memristances of memristors) will be fixed for
theoretically infinite period of time without a need for being
refreshed.

Note that for MISO system with N inputs, N memristor crossbars are
needed as N 2-dimensional discrete planes in IDS method.  After
dropping all of the training input-output data onto these memristor
crossbars, now they can be used in inferential process in IDS
method.
\subsection{Extraction of narrow path and spread values from pattern
images stored in memristor crossbars}

In ALM, with the incoming of any new input $\mathbf{x}^0$, inference
is done only through the values of narrow path and spread functions
at this point which have been computed by IDS method. As a result,
totally computation and storing of narrow path and spread functions
for every planes are not necessary. Instead, for any input data,
computation of the values of these two functions for each of the
input-output at this point planes will be sufficient for doing
inference. Fig. \ref{fig3} shows the circuit that we have proposed
for doing these computations. The inputs of this circuit, {\it i.e.}
$c_i$ for $i=1, 2, \ldots, m$ where m is the number of rows of the
crossbar, are connected directly to the horizontal wires of the
memristor crossbar. However, for having better understanding and
because the overall circuit was too much large, we have split the
whole circuit into two parts shown in Figs. \ref{fig1} and
\ref{fig3}. The crossbar in this figure (Fig. \ref{fig3}) is the
same as the previously described memristor crossbar but by this
difference that in those crosspoints which are specified by black
dots, instead of memristor, resistive junctions with a resistance
$R_{res}$ are being fabricated. In other crosspoints, there is no
junction between two perpendicular wires. As can be seen in Fig.
\ref{fig3}, each column of this resistive crossbar is connected to
an opamp having resistor with resistance value $R_{res}$ as a
feedback (except the last one which its feedback resistor is
$\frac{R_{res}}{2}$). Note that the combination of resistors of the
resistive crossbar and these opamps creates a opamp-based summing
circuits. In this case, outputs of these opamps denoted by $g_i$ for
$i=1, 2, \ldots, n$ can be written as:

\begin{figure}[!t]
\centering %\subfigure[Input recto image with show-through.]
{
% label for first subfigure
\fbox{\includegraphics[width=3.3in,height=2.8in]{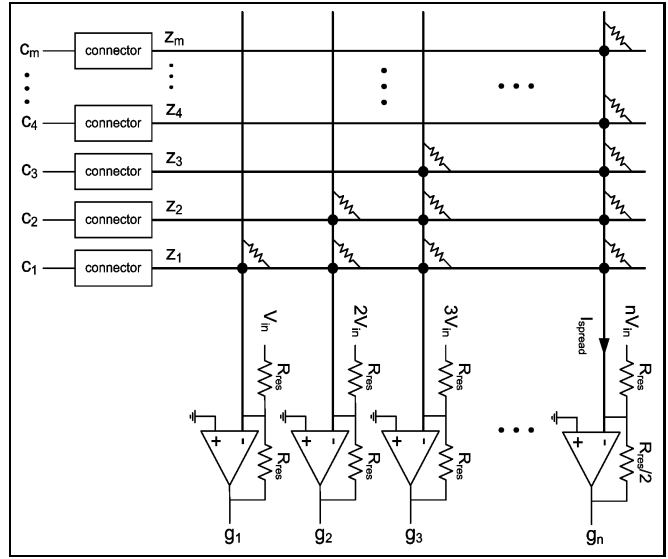}}}
\caption{Proposed circuit for computing the values of narrow path
and spread functions.}
\label{fig3} %% label for entire figure
\end{figure}

\begin{equation}\label{eq10}
g_i= \left\{
\begin{array}{rl}
\displaystyle{-\left[\sum_{t=1}^iz_t+i\times V_{in}\right]} &\quad
1\leq i<n,\\
\\\displaystyle{-\frac{1}{2}\left[\sum_{t=1}^nz_t+n\times V_{in}\right]}
&\quad i=n,
\end{array} \right.
\end{equation}
where $V_{in}$ is a constant positive value which will be defined in
the next paragraph and n=m is the number of columns in resistive
crossbar. Internal circuit of the {\it connector} blocks used in
Fig. \ref{fig3} and connecting memristor and resistive crossbars is
depicted in Fig. \ref{fig4:a}. To compute narrow path, these blocks
are configured as a simple current to voltage convertor (such as the
one shown in Fig. \ref{fig4:b}) by setting {\it clk2} input pin in
Fig. \ref{fig4:a} into logic 1. On the other hand, for the
calculation of the spread value, {\it connector} blocks are
configured as a circuit shown in Fig. \ref{fig4:c} by setting {\it
clk2} input pin into logic 0. This circuit converts the input
current I into voltage through the resistor $R_x$ and then compares
it with the pre-specified value $V_{th}$. Therefore, this circuit
can be used for thresholding.

Now, let's see how our proposed circuit can calculate narrow path
and spread functions from the pattern image stored in memristor
crossbar. For computing narrow path in this plane (memristor
crossbar) at any specific point $x_0$, the {\it j}th column of the
memristor crossbar which corresponds to $x_0$ should be connected to
the small positive voltage $V_{in}$ while other columns and rows are
connected into high impedance. Note that in this case, {\it clk1}
input pin of the memristor crossbar should be set into logic zero.
Smallness of the applied voltage is necessary for not changing the
memristances of the memristors while computing a narrow path.
Consequently, the output of the {\it i}th {\it connector} block
configured as a current-to-voltage converter will be:

\begin{figure}[!t]
\centering \subfigure[]{
\label{fig4:a} %% label for first subfigure
\includegraphics[width=2.5in,height=1.8in]{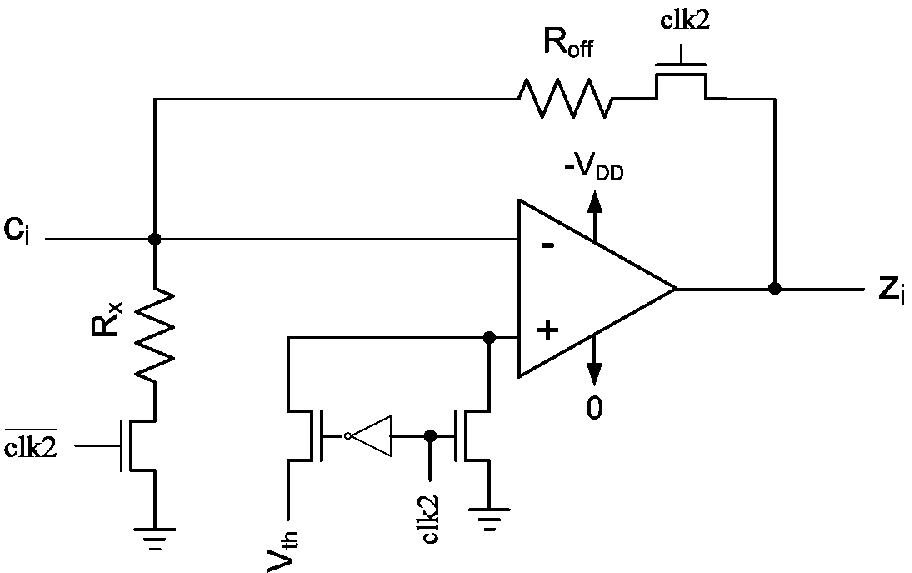}}
\hspace{0.05in} \subfigure[]{
\label{fig4:b} %% label for second subfigure
\includegraphics[width=1.2in,height=1.1in]{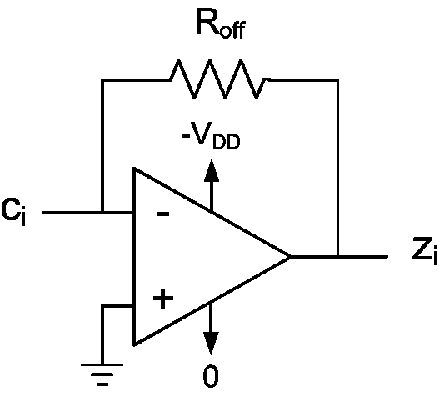}}
\vspace{0.05in} \subfigure[]{
\label{fig4:c} %% label for second subfigure
\includegraphics[width=1.5in,height=0.9in]{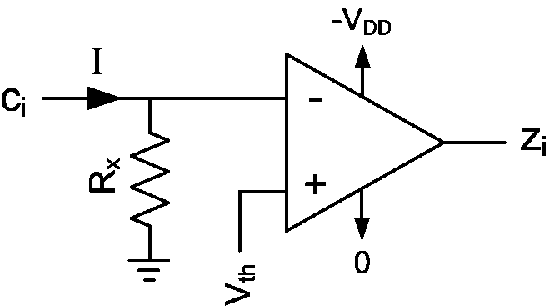}}
 \caption{(a) Internal circuit of the connector block. (b) Connector block
 configured as a current-to-voltage convertor. (c) Connector block configured
 for performing thresholding operation.}
\label{fig4} %% label for entire figure
\end{figure}

\begin{equation}\label{eq11}
z_i=-\frac{R_{off}}{R_{ij}}V_{in}\qquad i=1, 2, \ldots, m,
\end{equation}
where $R_{ij}$ is the memristance of the memristor at the crossing
point of the {\it i}th row and {\it j}th column of the memristor
crossbar after the ink dropping procedure. Equation \eqref{eq11} can
be rewritten as:

\begin{equation}\label{eq12}
z_i=-\frac{R_{off}}{R_{off}-\Delta R_{ij}}V_{in}\qquad i=1, 2,
\ldots, m,
\end{equation}
where $\Delta R_{ij}$ is the amount of change made to the
memristance of the memristor located at coordinate (i, j) during the
ink dropping process. As explained in previous section, since we
have assumed that $R_{off}$ is much higher than $\Delta R_{ij}$
({\it i.e.} $R_{off}\gg\Delta R_{ij}$), \eqref{eq12} can be
simplified as:

\begin{equation}\label{eq12}
z_i=-\frac{V_{in}}{1-\frac{\Delta
R_{ij}}{R_{off}}}\simeq-V_{in}\left (1+\frac{\Delta
R_{ij}}{R_{off}}\right )\qquad i=1, 2, \ldots, m.
\end{equation}

By substituting \eqref{eq12} into \eqref{eq10} we will have:

\begin{equation}\label{eq13}
g_i= \left\{
\begin{array}{rl}
\displaystyle{\frac{V_{in}}{R_{off}}\sum_{t=1}^i\Delta R_{tj}}
&\quad 1\leq i<n,\\ \\
\displaystyle{\frac{V_{in}}{2R_{off}}\sum_{t=1}^n\Delta R_{tj}}
&\quad i=n.
\end{array} \right.
\end{equation}

Tarkhan  {\it et al.} in \cite{tarkhan} have proved that if $b^*$ be
the index of the narrow path cell in the {\it j}th column of the
crossbar, it will has the following interesting property:

\begin{equation}
\sum_{t=1}^{b^*-1}\Delta R_{tj}<\frac{1}{2}\sum_{t=1}^{n}\Delta
R_{tj}\leq\sum_{t=1}^{b^*}\Delta R_{tj}.
\end{equation}

By multiplying sides of this inequality by $\frac{V_{in}}{R_{off}}$,
we will get:

\begin{equation}
\frac{V_{in}}{R_{off}}\sum_{t=1}^{b^*-1}\Delta
R_{tj}<\frac{V_{in}}{2R_{off}}{\sum_{t=1}^{n}\Delta
R_{tj}}\leq\frac{V_{in}}{R_{off}}\sum_{t=1}^{b^*}\Delta R_{tj},
\end{equation}
which is equal to the following inequality:

\begin{equation}
g_{b^*-1}<g_n\leq g_{b^*}.
\end{equation}

\begin{figure}[!t]
\centering %\subfigure[Input recto image with show-through.]
{
% label for first subfigure
\fbox{\includegraphics[width=3.3in,height=1.9in]{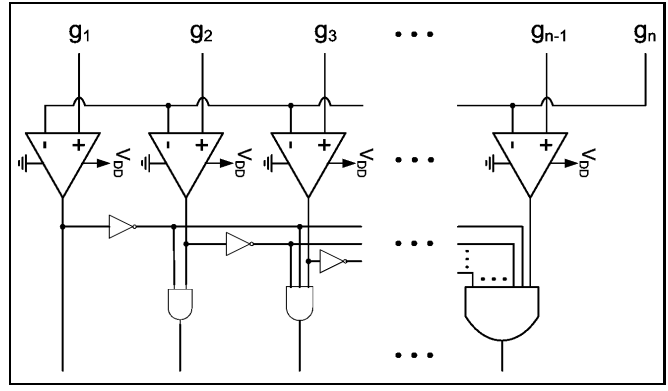}}}
\caption{Proposed circuit for finding the place of the narrow path
cell.}
\label{fig5} %% label for entire figure
\end{figure}

Therefore, the index of the narrow path cell can be simply found for
instance by using the circuit depicted in Fig. \ref{fig5}. In this
circuit, each opamp compares the input voltage $g_i$ with $g_n$.
Afterward, the combinational logic circuit specifies the first place
that the input signal $g_i$ becomes higher than $g_n$. Therefore,
during the narrow path calculation phase, only one of the outputs of
this combinational logic circuit will be at logic 1 which indeed
indicates the location of narrow path cell.

Finding the value of the spread function at $x_0$ can be done from
the same circuit shown in Fig. \ref{fig3}. If the {\it j}th column
of the memristor crossbar which corresponds to $x_0$ is connected to
the small negative voltage like $-V_{in}$ (while other wires are
connected to high impedance), then the voltage at the positive input
terminal of the opamp in the {\it i}th {\it connector} block (which
is configured as the circuit shown in Fig. \ref{fig4:c}) will be:

\begin{equation}
V_+^i=\frac{-V_{in}}{R_off-\Delta R_{ij}+R_x}\times R_x.
\end{equation}

By setting $V_{th}^i$ (the voltage at the negative input terminal of
the opamp in the {\it i}th {\it connector} block) to
$\frac{-V_{in}}{R_off-\Delta+R_x}\times R_x$ where $\Delta$ is a
predetermined constant value, output voltage of the {\it i}th {\it
connector} block, $z_i$, can be written as (see Fig. \ref{fig4:c}):

\begin{equation}\label{eq14}
z_i= \left\{
\begin{array}{rl}
-V_{DD} \qquad &\Delta R_{ij}\geq\Delta,\\ \\ 0 \qquad \ \ \ \
&\Delta R_{ij}<\Delta.
\end{array} \right.
\end{equation}

As we have stated at the beginning of this section, memristance of
all of the memristors change during the ink dropping procedure. To
calculate the spread function at $x_0$, it is necessary to determine
how many memristors on the {\it j}th column of the memristor
crossbar are being affected considerably enough during the ink
dropping process (see how the value of spread is determined in Fig.
\ref{fig9} at input point $x_i^0$). In \eqref{eq14}, $\Delta$ acts
as a threshold. If $\Delta R_{ij}$ (which represent the amount of
ink dropped to the coordinate (i, j)) be higher than $\Delta$, the
output of the {\it i}th {\it connector} block will be $-V_{DD}$.
Otherwise, the output will be zero. Since suitable value of $\Delta$
does not require to be precise and in addition, it depends on so
many factors such as $V_0$, $t_0$ and the number of training data,
it can be simply estimated for example by trial and error.
Therefore, the value of the spread at $x_0$ can be approximated by
the number of opamps (hereafter denoted by M) which their outputs
are $-V_{DD}$. This quantity can be computed through the current
passing from the last vertical wire of the resistive crossbar,
$I_{\text{spread}}$, which is shown in Fig. \ref{fig3} and can be
written as:

\begin{equation}
I_{spread}=-M\frac{V_{DD}}{R_{res}}.
\end{equation}

Consequently, this current is directly proportional to M and
therefore can be considered as a value of the spread function at the
presented input data. Hence, this proposed circuit, at any input
data, can calculate the values of narrow path and spread functions
in only two clock periods.

\section{Simulation results}
\label{simulation}

\begin{figure}[!t]
\centering %\subfigure[Input recto image with show-through.]
{
% label for first subfigure
\fbox{\includegraphics[width=3.3in,height=2.2in]{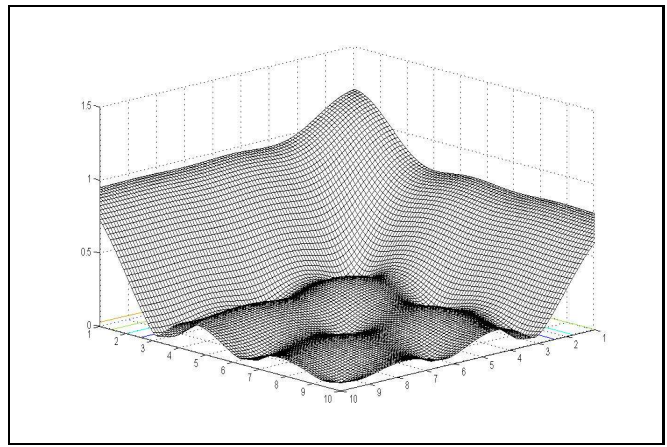}}}
\caption{Graph of function \eqref{eq16}.}
\label{fig6} %% label for entire figure
\end{figure}

For investigating the accuracy of the proposed method, we conducted
a modeling simulation. In this modeling test, the following
nonlinear function is used:

\begin{equation}\label{eq16}
y=\sqrt{2\left( \frac{\sin x_1}{x_1}\right)^2+3\left( \frac{\sin
x_2}{x_2}\right)^2},\qquad 1\leq x_1, x_2\leq10,
\end{equation}
where its graph is shown in Fig. \ref{fig6}. 800 randomly chosen
data sets are used for this modeling test and each $x_i-y$ planes of
IDS (such as the typical one shown in Fig. \ref{fig1}) has the
resolution of 100x90. This means that each of these crossbars has
9000 memristors. Resistance of the resistors which are connecting
rows or columns to each other in these memristor crossbars was set
to 1000$\Omega$. These planes are simulated in HSPICE software by
utilizing the SPICE model proposed in \cite{Biolek} for memristors.

\begin{figure}[!t]
\centering %\subfigure[Input recto image with show-through.]
{
% label for first subfigure
\fbox{\includegraphics[width=3.3in,height=2in]{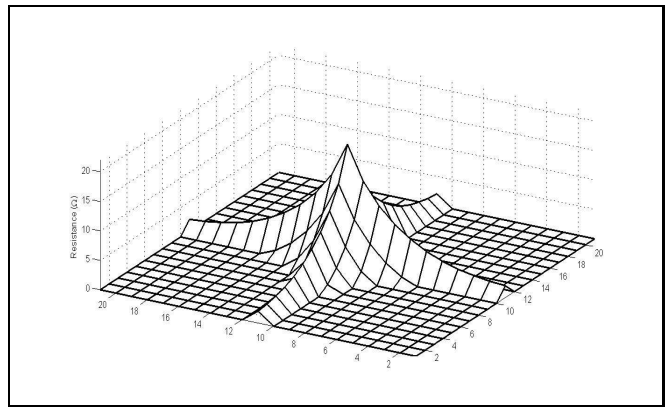}}}
\caption{Absolute value of the amount of changes made to the
memristance of the memristors near to the distillation point of an
ink drop.}
\label{fig7} %% label for entire figure
\end{figure}

\begin{figure}[!t]
\centering \subfigure[]{
\label{fig10:a} %% label for first subfigure
\includegraphics[width=3.7in,height=2.4in]{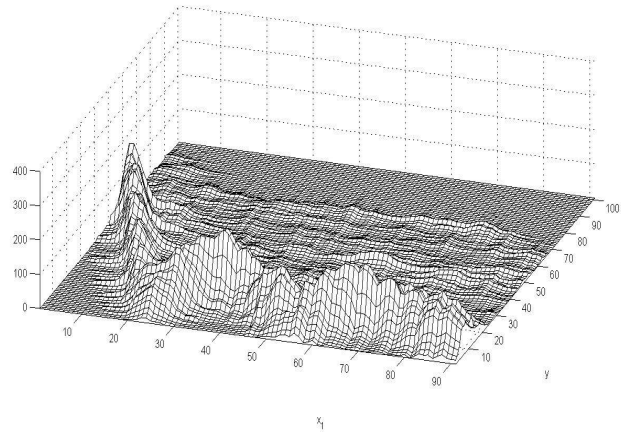}}
\hspace{0.05in} \subfigure[]{
\label{fig10:b} %% label for second subfigure
\includegraphics[width=3.7in,height=2.4in]{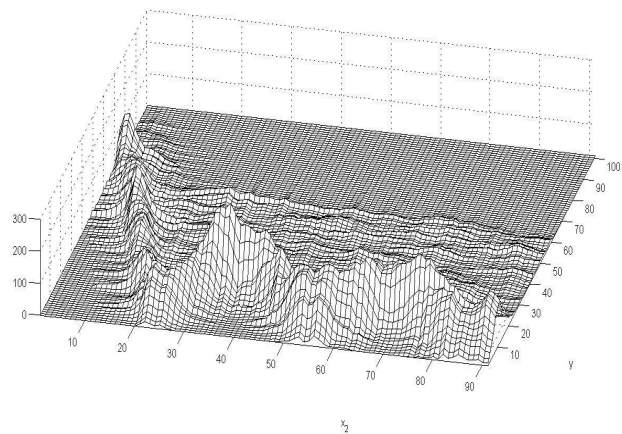}}
 \caption{Results of spreading all of 800 training
data sets on $x_1-y$ and $x_2-y$ planes respectively.}
\label{fig10} %% label for entire figure
\end{figure}

\begin{figure}[!t]
\centering \subfigure[]{
\label{fig11:a} %% label for first subfigure
\includegraphics[width=2in,height=1.3in]{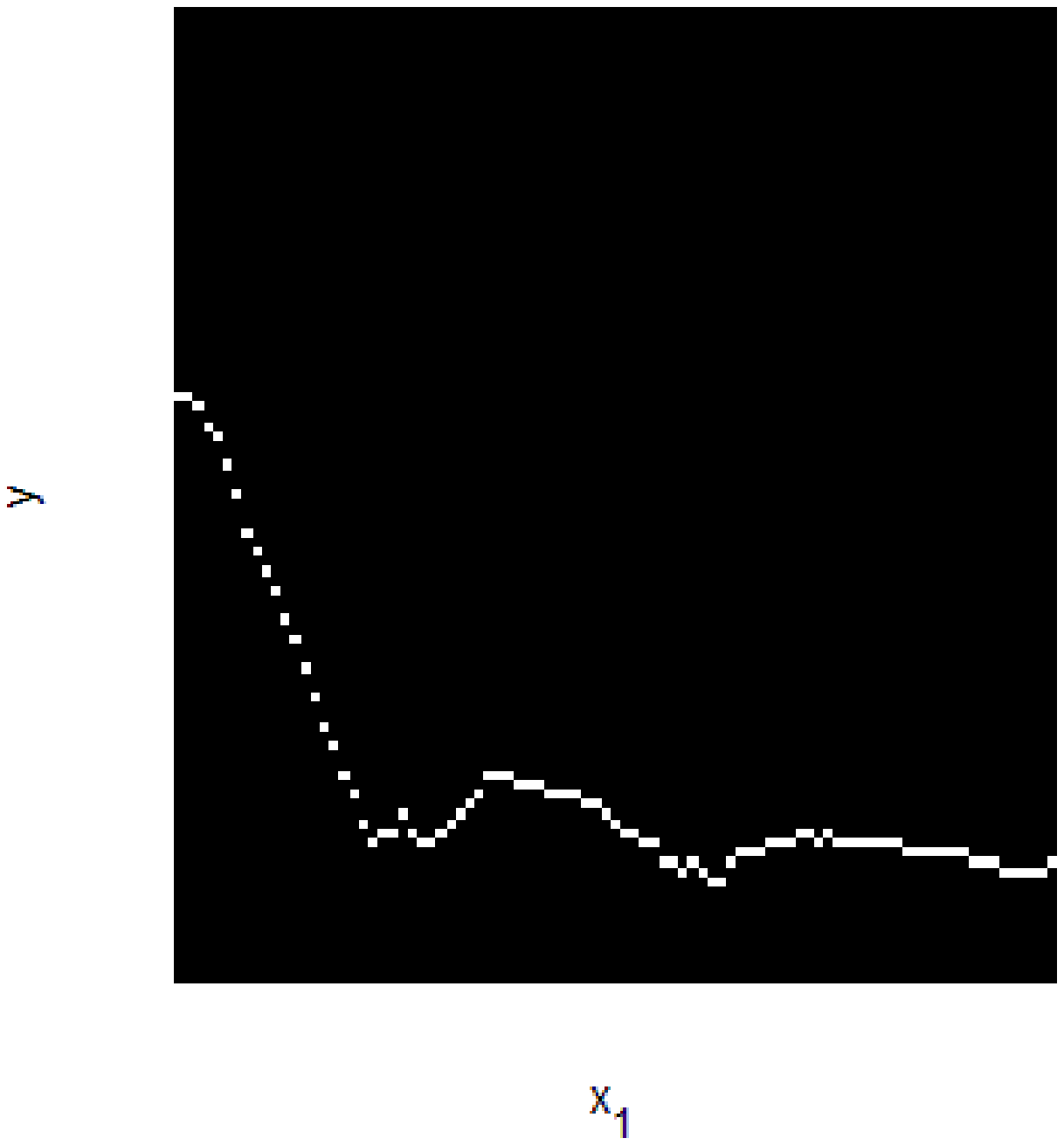}}
\hspace{0.2in} \subfigure[]{
\label{fig11:b} %% label for second subfigure
\includegraphics[width=2in,height=1.3in]{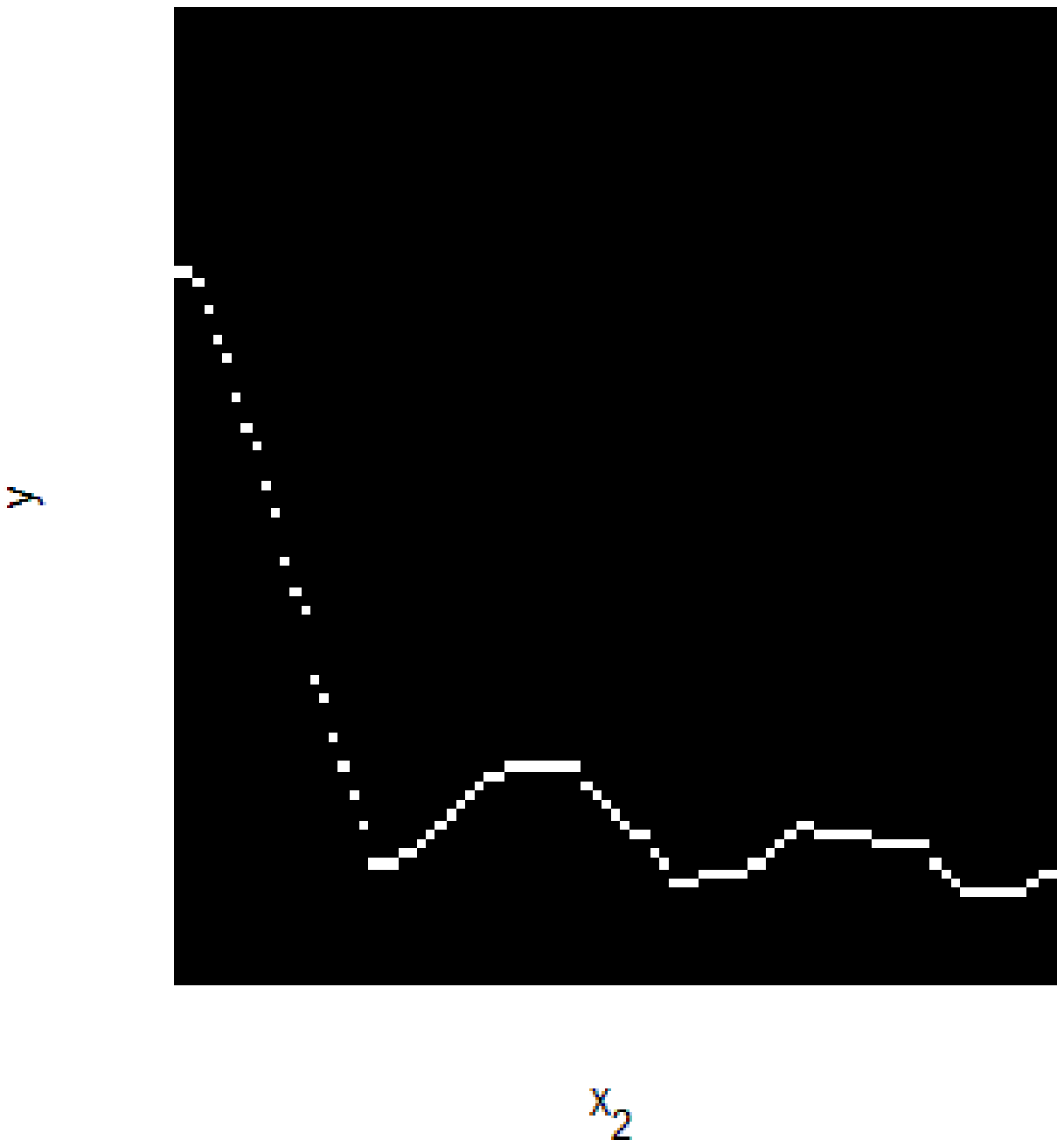}}
 \caption{Computed narrow paths from the $x_1-y$ and $x_2-y$ planes shown in Fig. \ref{fig10}.}
\label{fig11} %% label for entire figure
\end{figure}

\begin{figure}[!t]
\centering \subfigure[]{
\label{fig12:a} %% label for first subfigure
\includegraphics[width=3.7in,height=1.6in]{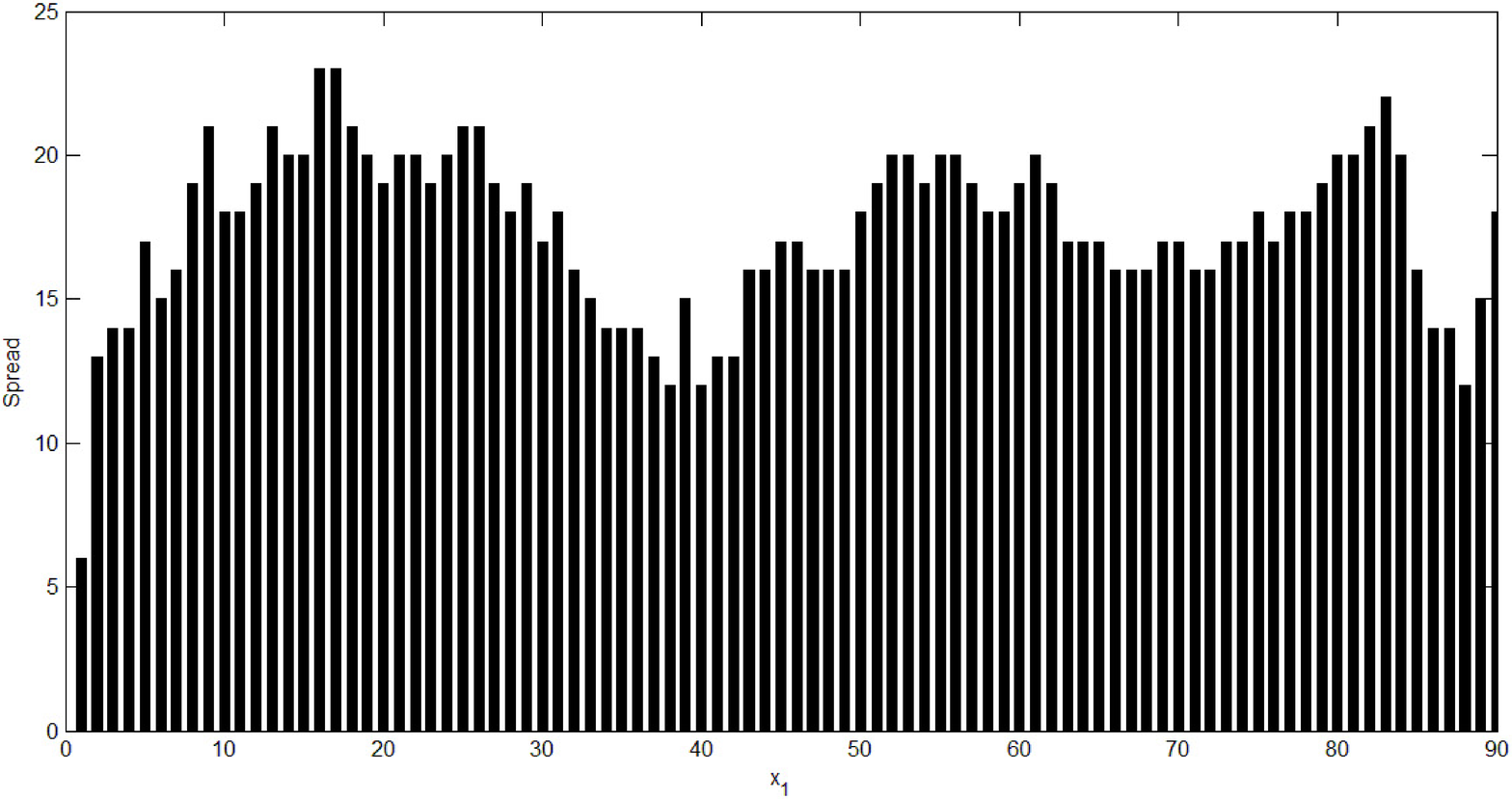}}
\hspace{0.05in} \subfigure[]{
\label{fig12:b} %% label for second subfigure
\includegraphics[width=3.7in,height=1.6in]{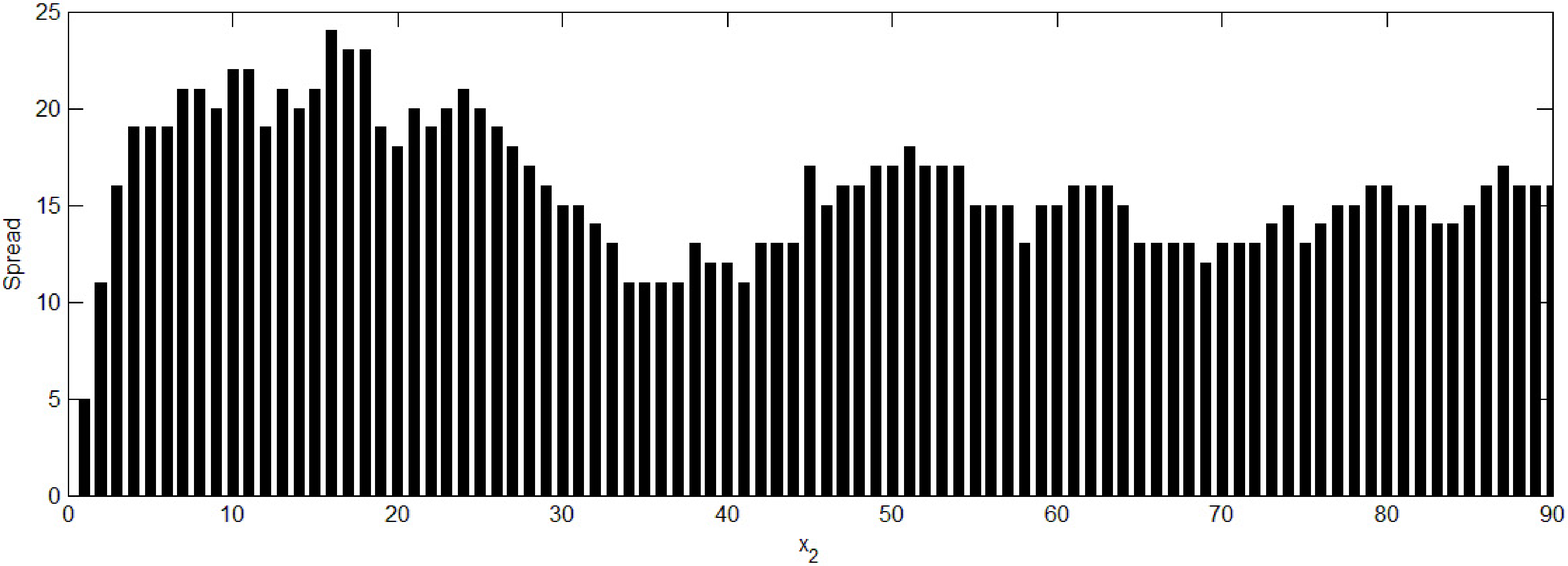}}
 \caption{Computed spreads from the $x_1-y$ and $x_2-y$ planes shown in Fig. \ref{fig10}.}
\label{fig12} %% label for entire figure
\end{figure}

\begin{figure}[!t]
\centering %\subfigure[Input recto image with show-through.]
{
% label for first subfigure
\fbox{\includegraphics[width=3.3in,height=2.2in]{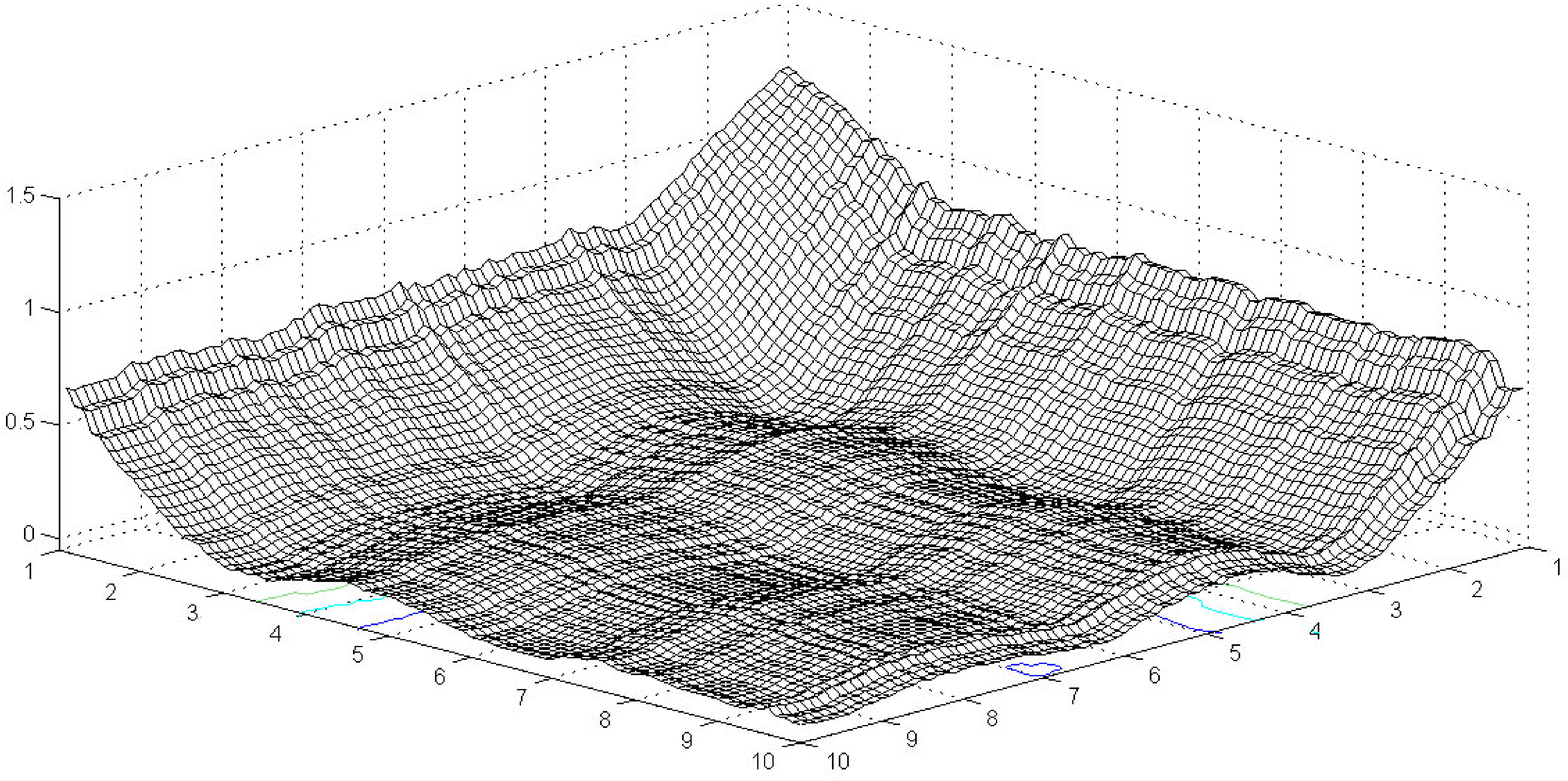}}}
\caption{Modeling result for \eqref{eq16}.}
\label{fig13} %% label for entire figure
\end{figure}

As explained in Section \ref{main-1}, distilling an ink drop at any
point changes the memristance of the neighboring memristors in the
crossbar. Figure \ref{fig7} shows the absolute value of the amount
of changes made to the memristance of the memristors near to the
distillation point of one ink drop. Note that this result is
obtained by setting $R_{off}$, $V_0$ and $t_0$ into 100k$\Omega$, -3
volt and 10 milliseconds respectively. Results of spreading all of
800 training data sets on $x_1-y$ and $x_2-y$ planes are shown in
Fig. \ref{fig10} (note that these figures show $\Delta R_{ij}$ for
any of the memristors in the crossbar) and their corresponding
computed narrow paths are depicted in Fig. \ref{fig11}. For the
calculation of the spread at any input point from the planes shown
in Fig. \ref{fig10}, we set $\Delta$ into 20$\Omega$. Results
computed by our proposed hardware for the values of spread ($M$) at
any column of the planes shown in Fig. \ref{fig10} are presented in
Fig. \ref{fig12}. Finally, by passing these computed values for
narrow path and spread functions to ALM, obtained modeling result
for \eqref{eq16} is shown in Fig. \ref{fig13}.

By comparing \ref{fig6} and \ref{fig13}, it is clear that our
proposed method can effectively be used as a hardware implementation
of IDS method.

\section{Conclusion}
\label{conclusion}

In this paper a new hardware implementation of IDS method is
presented based on the memristor crossbar structure. Proposed
structure can compute narrow path and spread at any given input data
in real-time since it is implemented in analog instead of digital.
Furthermore, since input-output training data are stored in
memristors at the crosspoints of the crossbar, system can continue
working properly after the occurrence of power breakdown. However,
since memristor can be realized in nanoscales, we confront some
difficulties and technological challenges in the integration of
nanoscale memristor crossbars and conventional electronics systems.

\end{document}